\documentclass[runningheads]{llncs}
\usepackage[T1]{fontenc}
\usepackage{graphicx}
\usepackage{amsmath} 
\usepackage{amssymb}
\usepackage{wrapfig}
\usepackage{siunitx}
\usepackage{booktabs} 
\begin{document}
%
\title{MsFormer: Enabling Robust Predictive Maintenance Services for Industrial Devices}
%
%
%

\author{Jiahui Zhou\inst{1} \and
Dan Li\inst{1}\thanks{Corresponding author.} \and
Ruibing Jin\inst{2} \and
Jian Lou\inst{1} \and
Yanran Zhao\inst{3} \and \\
Zhenghua Chen\inst{2} \and
Zigui Jiang\inst{1} \and
See-Kiong Ng\inst{4}}

\authorrunning{J. Zhou et al.}

%
%


\institute{School of Software Engineering, Sun Yat-Sen University, China \\
\and
Agency for Science, Technology and Research (A*STAR), Singapore \\
\and
School of the Art Institute of Chicago, Chicago \\
\and
Institute of Data Science and School of Computing, \\National University of Singapore, Singapore \\
}

\maketitle              
\begin{abstract}

Providing reliable predictive maintenance is a critical industrial AI service essential for ensuring the high availability of manufacturing devices. Existing deep-learning methods present competitive results on such tasks but lack a general service-oriented framework to capture complex dependencies in industrial IoT sensor data. While Transformer-based models show strong sequence modeling capabilities, their direct deployment as robust AI services faces significant bottlenecks. Specifically, streaming sensor data collected in real-world service environments often exhibits multi-scale temporal correlations driven by machine working principles. Besides, the datasets available for training time-to-failure predictive services are typically limited in size. These issues pose significant challenges for directly applying existing models as robust predictive services. To address these challenges, we propose MsFormer, a lightweight Multi-scale Transformer designed as a unified AI service model for reliable industrial predictive maintenance. MsFormer incorporates a Multi-scale Sampling (MS) module and a tailored position encoding mechanism to capture sequential correlations across multi-streaming service data. Additionally, to accommodate data-scarce service environments, MsFormer adopts a lightweight attention mechanism with straightforward pooling operations instead of self-attention. Extensive experiments on real-world datasets demonstrate that the proposed framework achieves significant performance improvements over state-of-the-art methods. Furthermore, MsFormer outperforms across industrial devices and operating conditions, demonstrating strong generalizability while maintaining a highly reliable Quality of Service (QoS).

\end{abstract}
\section{Introduction}
\label{sec:introduction}

Degradation and failure of critical industrial equipment can lead to severe service outages and economic loss. To ensure high availability and maintain a reliable Quality of Service (QoS), Prognostics and Health Management (PHM) mitigates such losses by building predictive models from sequential sensor data from industrial devices to forecast a machine's remaining useful time until failure (RUL) and guide optimal maintenance scheduling. With the paradigm shift towards AI-as-a-Service (AIaaS) and the prosperity of related applications, deep-learning methods have been widely deployed as data-driven predictive services for industrial assets \cite{Robotics}. Existing CNN-based service models \cite{jin2022position} employ stacked convolutional layers for multi-dimensional IoT sensor feature extraction. While CNNs capture spatial context via parallel processing, they insufficiently model long-range temporal patterns in streaming data. Conversely, RNN-based methods handle streaming sensor data via hidden states, but their sequential recurrence leads to computational burden and high latency, making them prone to vanishing gradients on long sequences \cite{nips_rnnvanishing}.

Recent years have witnessed growing interest in adapting the Transformer architectures to analyze sequential sensor measurements for industrial devices. Transformers can be trained in parallel and flexibly capture the relationships among tokens within the input sequence, thereby avoiding the gradient vanishing and computationally slow issues \cite{pascanu2013difficulty}. Prominent models such as FEDformer \cite{zhou2022fedformer}, Autoformer \cite{wu2021autoformer}, and iTransformer \cite{liu2024itransformer} have demonstrated effectiveness in general time series forecasting and motivated efforts in building time series foundation models with modified Transformer structures \cite{Timexer}. Transformer-based approaches have also been adapted to the time-to-failure predictive services\cite{gao2025multiscale,gao2024nonlinear,liang2023remaining,wang2024dvgtformer}. Despite the primary success, researchers have identified bottlenecks with directly applying the existing Transformer architecture to time series prediction tasks \cite{wang2024timemixer,lu2024incontextpredictor,Zeng2022AreTE}.

Specifically, the self-attention mechanism adopted by existing Transformer-based models cannot adequately capture the complex long-term degradation patterns via direct consecutive timestamps-to-tokens conversion of streaming IoT service data \cite{liu2025timerxl,masserano2025enhancing}. Unlike language tokens, which are sequentially correlated by semantics and syntax, industrial sensor measurements are often semantically sparse across operating cycles \cite{liu2024itransformer,wu2021autoformer} and exhibit multi-scale dependencies across temporal resolutions, rendering the vanilla Transformer less efficient \cite{Zeng2022AreTE}. This multi-scale structure aligns with the intuition that the health status of devices at a given working cycle is influenced by both nearby and distant operation conditions. Moreover, the equipment-specific datasets for training time-to-failure predictive services are often limited in size, rendering existing Transformer-based architectures with stacked dense network modules and large network depth perform poorly and fail to guarantee a reliable QoS in data-scarce environments. Thus, capturing multi-scale temporal dependencies and complex degradation patterns among streaming sensor measurements and designing a lightweight, service-oriented architecture tailored to limited industrial datasets remain the key challenges for the successful application of Transformers as a unified AI service model for reliable industrial predictive maintenance.

In this work, we propose MsFormer, a lightweight Multi-scale Transformer designed as a unified AI service model for reliable industrial predictive maintenance. The Multi-scale Sampling (MS) module is developed to restructure the original timestamps for multiple time horizons, allowing MsFormer to model temporal dependencies at different scales. We further introduce a multi-scale positional encoding that assigns learned weights to relative positions after sampling to better capture cross-scale correlations within the multi-streaming service data. Besides, MsFormer employs a lightweight attention mechanism that employs straightforward pooling operations instead of self-attention, tailored to the smaller-scale industrial datasets. In summary, our contributions are as follows:

\begin{enumerate}
\item We propose a four-stage framework for unified time-to-failure prediction. In the first two stages, the proposed method deals with the original sensor measurement data with the Multi-scale Sampling (MS) module and the lightweight attention mechanism. In the last two stages, the encoded embeddings are further dealt with by the MS module with multi-scale position encoding. Sampling scales are differently distributed to adequately grasp the multi-scale temporal correlations essential for reliable predictive services.  

\item The proposed MS module is dedicated to capturing intricate multi-scale temporal correlations across various timestamps effectively through restructuring the original timestamps for multiple time horizons. The multi-scale position encoding mechanism is proposed to enhance MsFormer's ability to extract cross-scale correlations among sensor measurements. The lightweight attention mechanism employs straightforward pooling operations instead of self-attention, enabling MsFormer to be more adaptable to smaller-scale industrial datasets.

\item We conduct extensive experiments on public datasets collected on diverse devices and operating conditions to validate MsFormer as a unified AI service model. Experimental results show that MsFormer surpasses state-of-the-art methods not only in prediction performance but also in computational efficiency. Ablation results also show the effectiveness of the modules we proposed in MsFormer.
\end{enumerate}


\vspace{-1.5em}
\section{Related Works}
\label{sec:RW}
\vspace{-0.5em}
\subsection{Transformer Models in Terms of Feature Extraction}
Existing studies introduce frequency-based feature extraction techniques to better capture the inherent characteristics of industrial service data. MLEAN \cite{liu2024multi} captures diverse temporal patterns through multi-scale and multi-frequency feature learning, while TFSCL \cite{jiang2025time} optimizes model provisioning for AI services through a time-frequency fusion contrastive loss function. Similarly, TATFA-Transformer \cite{zhang2024trend} incorporates a Bi-GRU as a trend augmentation module to capture latent temporal dependencies. Besides, some studies attempt to incorporate spatial correlations to capture the inherent characteristics of time series data. DCFA \cite{gao2023dual} applies dual-branch attention mechanism to independently weight temporal and spatial features, whereas GAT-DAT \cite{liang2023remaining} fuses node features to capture spatial correlations. 
However, these methods primarily focus on long-term dependencies across whole sequences. Given the intertwined and overlapping fluctuations within real-world service environments, inter-temporal relationships can vary across different time scales, potentially introducing biases into temporal dependency modeling.

\vspace{-1.0em}
\subsection{Transformer Models in Terms of Tailored Architectures}
As for designing tailored architectures to adapt the Transformer to time series analysis, TF-SCN \cite{jing2022transformer} employs a transformer-based backbone to compress features into a hierarchical latent space. Recent studies have explored graph transformer approaches. NSD-TGTN \cite{gao2024nonlinear} proposes a slow-varying dynamics-assisted temporal graph transformer, while DVGTformer \cite{wang2024dvgtformer} introduces a dual-view graph transformer that effectively integrates temporal and spatial information. Similarly, DS-STFN \cite{zhang2024dual} leverages a temporal convolutional network to extract features from grid-structured data, while employing a graph transformer module to capture spatial features from graph-structured data. However, these methods are not well-suited as reliable AI services for time-to-failure prediction, as the limited scale of training data is insufficient to effectively train a graph transformer. To address the issue that stacking network layers often fails to improve model performance, PBMT \cite{fang2024pbmt} employs a probsparse self-attention mechanism to reduce computational complexity. Meanwhile, IMDSSN \cite{zhang2023integrated} integrates both probsparse and logsparse self-attention mechanisms.

\begin{figure}[htbp]
\vspace{-2.0em}
  \centering
  \includegraphics[width=0.8\linewidth]{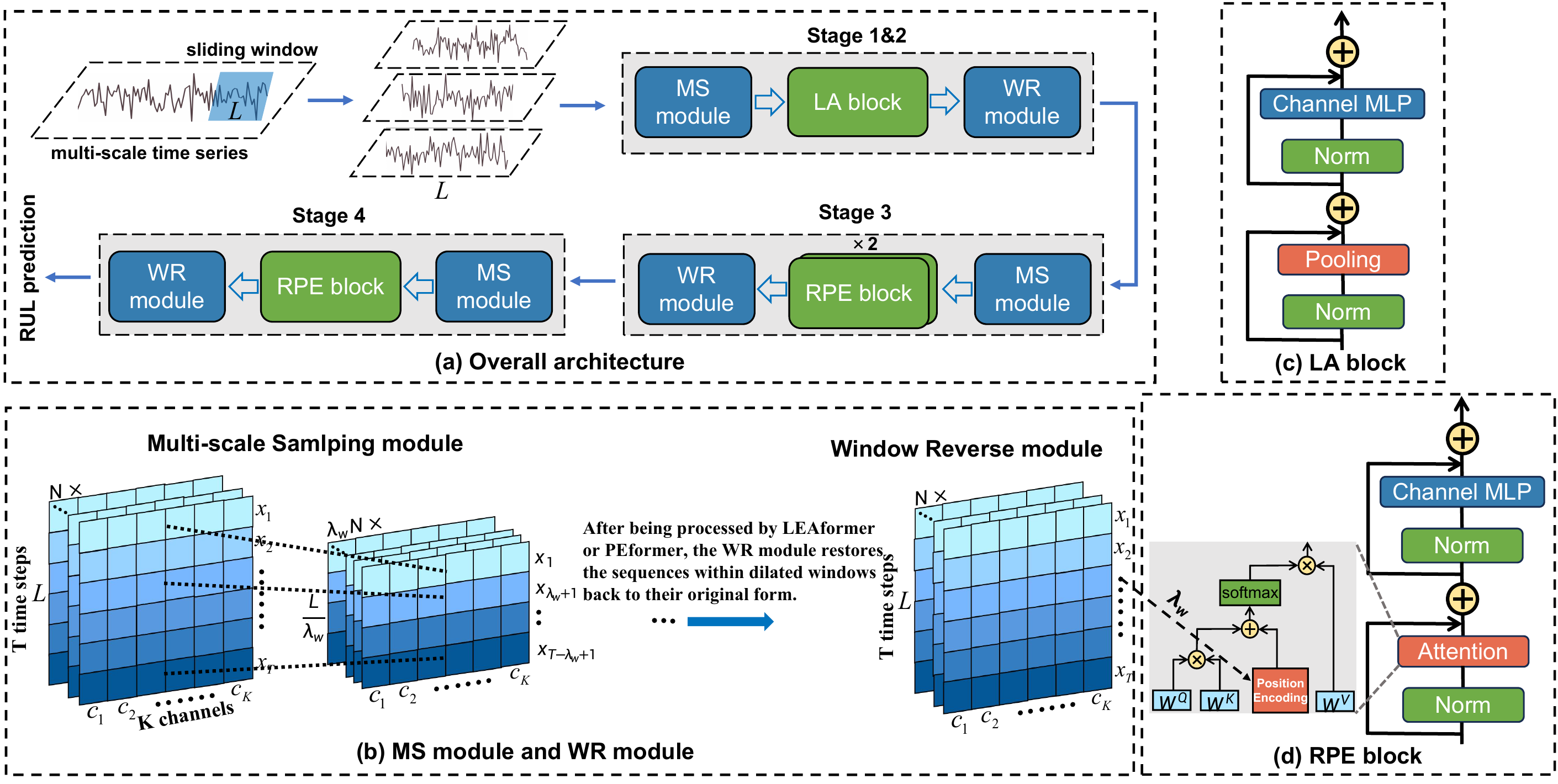}
  \caption{The overall architecture of the MsFormer.}
  \label{fig_overall_architecture}
  \vspace{-3.0em}
\end{figure}

\vspace{-0.5em}
\section{Methodology}
\label{sec:methodology}
\vspace{-0.5em}
\subsection{Overall Architecture}

The proposed MsFormer is a lightweight, hierarchical service-oriented framework for reliable industrial time-to-failure prediction. As illustrated in Figure \ref{fig_overall_architecture}, the input streaming service data is preprocessed using a sliding window of length ${L}$ and sequentially passed through four stages. In the first two stages, the Multi-scale Sampling (MS) module, Lightweight Attention (LA) block, and Window Reverse (WR) module are combined to capture temporal correlations at multiple resolutions. The MS module restructures sub-sequences at varying dilation rates to generate down-sampled sequences, preserving temporal order, while the LA block processes smaller-scale data using pooling operations instead of self-attention. The WR module then restores the sequence to its original resolution for the next stage.

In the last two stages, the MS module is paired with a Relative Position Encoding (RPE) block and the WR module to further enhance the robustness of the predictive service. The RPE block employs a multi-scale position encoding mechanism and self-attention to capture temporal dependencies, incorporating the down-sampling factor $\lambda_w$ to represent time intervals. This hierarchical design enables MsFormer to efficiently capture complex temporal patterns across multiple resolutions while maintaining computational efficiency.

\vspace{-1.0em}
\subsection{Multi-scale Sampling Module}
The gradual degradation of industrial equipment might obscure changes between proximate timestamps, making it difficult for dense attention mechanisms to capture critical degradation patterns essential for reliable predictive services\cite{zhou2022fedformer,liu2022pyraformer}. While most Transformer-based models focus on long-term dependencies \cite{gao2024nonlinear,gao2025multiscale}, time series data often exhibit varying dependency patterns across resolutions. The proposed MS module mitigates this by segmenting data into down-sampled sequences with different sampling rates, thus enhancing semantic richness and capturing temporal correlations driven by machine working principles. Specifically, it samples fixed-length patches with a down-sampling factor \( \lambda_w = {L}/{W} \), where ${L}$ is the patch length, and ${W}$ is the length of the down-sampled sequences. 

Then, new down-sampled sequences are generated via restructuring the original data points with different sampling rates while maintaining their temporal order as in the original sequence, where the temporal intervals across adjacent timestamps are reflected by the down-sampling factor:
\begin{small}
\begin{equation}
X^{(i)} = \{x_j | j = k\lambda_w + i, \, k = 0, 1, \ldots, \left\lfloor {L}/{\lambda_w} \right\rfloor - 1 \}
\end{equation}
\end{small}
where \( X \in \mathbb{R}^{N \times L \times C} \) represents the raw data,  \(X^{\text{(i)}}\) denotes the \( i \)-th down-sampled sequence, and \( x_j \) represents the data point at the \( j \)-th timestamp. Ultimately, the down-sampled sequences are reassembled along the $0$-th dimension:
\begin{small}
\begin{equation}
X_{\textit{dilated}} = \textit{concat}\left(X^{(0)}, X^{(1)}, \ldots, X^{(\lambda_w - 1)}\right)
\end{equation}
\end{small}

MsFormer addresses information redundancy and pattern complexity through a progressive strategy that adaptively adjusts the down-sampling factor \( \lambda_w \) in each stage to capture multi-scale temporal correlations. Specifically, in the first two stages, the down-sampling factor is set as 4. This maximizes the temporal gap between data points, thus effectively mitigating the sparsity in the correlation among adjacent data points. At the third stage, \( \lambda_w \) is set as 2, thereby extending the down-sampled sequence to half of the length of the original sub-sequence. This enhancement, coupled with the self-attention mechanism and the position encoding from the RPE block, enables MsFormer to delve deeper into the intrinsic temporal dependencies while remaining adaptable to data-scarce service environments. Subsequently, after efficiently mitigating redundancy through prior stages, the window length is recovered to the original sub-sequence length for direct feature learning.

\subsection{Structure of Transformer Blocks}
Following the architecture of the Vanilla Transformer, the input sequence of the Transformer blocks within MsFormer is initially encoded as follows:
\begin{small}
\begin{align}
{X_e} & = \text{InputEmb}({X}) \label{eq:InputEmb}
\end{align}
\end{small}
where \( X_e \) represents the embedding tokens, characterized by the dimensions \( \mathbb{R}^{N \times L \times C} \), with \( N \) indicating the batch size, \( L \) indicating the sequence length, and \( C \) is the encoding dimension. Subsequently, the embedding tokens are fed into the repeated Transformer blocks. Specifically, within each block, the following operations are performed:
\begin{small}
\begin{equation}
\begin{aligned}
\hat{X}_m^l = \text{TokenMixer}(\text{Norm}(X_e^{l-1})) + X_e^{l-1}, \\
X_m^l = \text{MLP}(\text{Norm}(\hat{X}_m^l)) + \hat{X}_m^l
\end{aligned}
\end{equation}
\end{small}
where \( \text{Norm}(\cdot) \) signifies normalization, and \( \text{TokenMixer}(\cdot) \) denotes the mechanism for mixing token information. \( \hat{X}_m^l \) and \( X_m^l \) respectively denote the intermediate and final feature representations within the \( l \)-th block.
\vspace{-1.0em}
\subsubsection{Lightweight Attention Mechainism}
The LA block is dedicated to streamlining feature abstraction and enhancing training efficiency for industrial time-to-failure prediction tasks. At the first two stages of the MsFormer, LA replaces the attention mechanism with a straightforward pooling operation. This design enhances MsFormer's performance and exhibits linear computational complexity in terms of sequence length as the token mixer.
\begin{small}
\begin{align}
\hat{X}_p^l = \text{Pooling}(\text{Norm}(X_e^{l-1})) + X_e^{l-1}
\end{align}
\end{small}
The pooling operation within the LA efficiently abstracts the essential temporal features from dilated windows, making it well-suited for smaller-scale industrial datasets while reducing overall model parameters.
\vspace{-1.0em}
\subsubsection{Relative Position Encoding Block}
The RPE block delves into the intricacies of spatial-temporal dependencies, which are implemented in the last two stages of the proposed MsFormer. The self-attention mechanism maps a query and a set of key-value pairs to an output. We denote \( x = (x_1, \ldots, x_n) \), where \( x_i \in \mathbb{R}^{d_x} \) represents the embedding vector processed by the MS module. The self-attention mechanism computes an output sequence \( z = (z_1, \ldots, z_n) \), where \( z_i \in \mathbb{R}^{d_z} \). Each output element \( z_i \) is calculated as a weighted sum of the input elements, and each weight coefficient \( \alpha_{ij} \) is computed by the softmax:
\begin{small}
\begin{align}
z_i = \sum_{j=1}^{n} \alpha_{ij} (x_j W^V), \quad \alpha_{ij} = \frac{\exp(e_{ij})}{\sum_{k=1}^{n} \exp(e_{ik})}
\end{align}
\end{small}
where \( e_{ij} \) is calculated by the scaled dot-product self-attention mechanism and augmented by our proposed multi-scale position encoding mechanism:
\begin{small}
\begin{align}
e_{ij} = \frac{(x_i W^Q) (x_j W^K)^T}{\sqrt{d_z}} + p_{ij}
\end{align}
\end{small}
To effectively distribute attention across various representational subspaces at distinct positions simultaneously, we employ ${h}$ parallel attention mechanisms, known as multi-head attention. The outputs of multiple attention heads are concatenated and linearly transformed to the desired dimensions.

\subsection{Multi-scale Position Encoding Mechanism}
\label{subsec:rpe}
The self-attention mechanism inherently struggles to capture the sequential nature of input tokens \cite{shaw-etal-2018-self,chen-etal-2021-simple}. Moreover, the MS module restructures time points across various patch sizes, resulting in a shift in the relative position of data points compared to their original sequence order. To mitigate these issues, we propose a multi-scale position encoding mechanism to enhance the temporal dependencies from a spatial perspective.

\begin{wrapfigure}{R}{0.5\textwidth}
    \vskip -1.5em
    \begin{minipage}{\linewidth}
    \centering 
    \includegraphics[width=\textwidth]{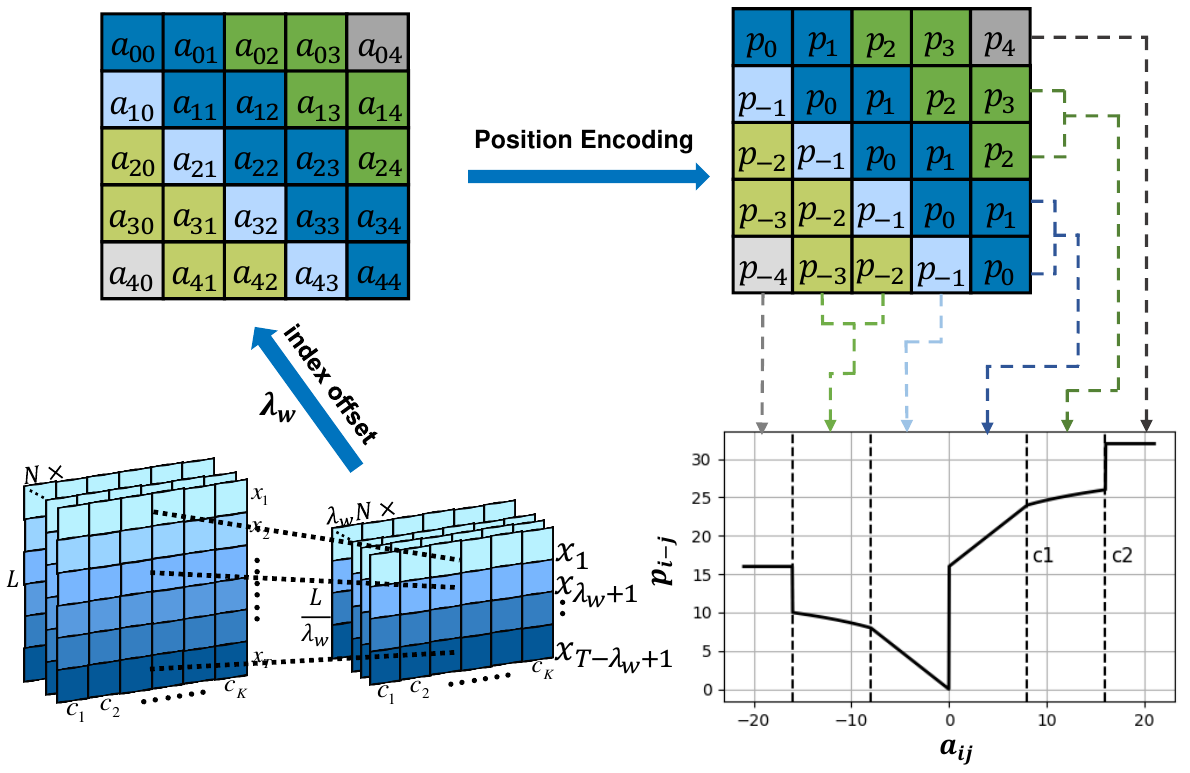} 
        \vskip -1.0em
    \caption{The illustration for the multi-scale position encoding mechanism.}
    \label{fig_index_func} 
        \vspace{-2.0em}
\end{minipage}
\end{wrapfigure}



While the relative positions of adjacent time steps remain consistent across sequences, their actual positions in the original sequence can vary significantly, as shown in Figure \ref{fig_index_func}. We construct a matrix \( A \in \mathbb{R}^{L \times C} \) to preliminarily describe the relative positions of elements in embedding vectors based on down-sampled sequences, where each element \( a_{ij} \) is defined as \(a_{ij} = (i-j) \times \lambda_w\). To reduce computation costs, we approximate the index offset \( {a}_{ij} \) by sharing positional parameters and define a position encoding matrix \( P \in \mathbb{R}^{L \times C} \), where \( L \)  is the sequence length and \( C \) is the number of channels.

We then employ a piecewise index function that maps the element \( {a}_{ij} \) to \(| p_{i-j} |\). Specifically, for \(|{a}_{ij}|\) smaller than \( c_1 \), precise encoding is applied due to the higher position encoding importance. When \(|{a}_{ij}|\) falls between \( c_1 \) and \( c_2 \)), elements with similar values share position encoding, with the encoding range expanding logarithmically as the offset increases. Finally, the index is clipped to the upper limit \( c_2 \) when \(|{a}_{ij}|\) exceeds this bound.

\begin{small}
\begin{align}
|p_{i-j}| = 
\begin{cases} 
\left| {a}_{ij} \right|, & \text{if } \left| {a}_{ij} \right| < c_1 \\
c_1 + c_1 \times \frac{\log(\left| {a}_{ij} \right| / c_1)}{\log(128 / c_1)}, & \text{if } c_1 \leq \left| {a}_{ij} \right| < c_2 \\
c_2, & \text{if } \left| {a}_{ij} \right| \geq c_2 
\end{cases}
\end{align}
\end{small}

Ultimately, to indicate the bidirectional nature of relative positions, for positive elements \( a_{ij} \) in matrix \( A \), we sum them with the offset \( c_2 \) to yield the final position encoding \( p_{ij} \). The values of \( c_1 \) and \( c_2 \) are determined based on the lengths of the original and down-sampled sequences.
\begin{small}
\begin{align}
p_{i-j} = 
  \begin{cases} 
   |p_{i-j}| + c_2, & \text{if } a_{ij} \geq 0 \\
   |p_{i-j}|, & \text{if } a_{ij} < 0 
  \end{cases}
\end{align}
\end{small}

\vspace{-1.5em}
\section{Experiments}
\label{sec:exp}

\begin{wraptable}{R}{0.6\textwidth} 
    \vspace{-3.5em} 
    \centering
    \small
    \caption{Details of the C-MAPSS Dataset.}
    \begin{tabular}{ccccc}
    \hline
    Subsets & FD001 & FD002 & FD003 & FD004 \\
    \hline
    Operation conditions & 1 & 6 & 1 & 6 \\
    Fault modes & 1 & 1 & 2 & 2 \\
    Training trajectories & 100 & 260 & 100 & 248 \\
    Testing trajectories & 100 & 259 & 100 & 249 \\
    \hline
    \end{tabular}
    \vspace{-2.5em} 
    \label{table:cmapss}
\end{wraptable}

\vspace{-0.5em}
\subsection{Dataset and Experimental Settings}
\subsubsection{Datasets}
The C-MAPSS archive \cite{saxena2008damage} summarized in Table \ref{table:cmapss} is a widely used benchmark for time-to-failure prediction. It records 21 engine sensor measurements and comprises four sub-datasets, each simulating different engine operating scenarios. Notably, FD002 and FD004 datasets are more complex than FD001 and FD003, involving six different operating conditions during the recording periods. To validate the generalization capability of the proposed MsFormer, we further conduct experiments on the NASA dataset \cite{chen2024attmoe}, which comprises records from four lithium-ion batteries. Each battery undergoes repeated charging and discharging cycles, with current, voltage, and impedance measured for each cycle. 


\vspace{-1.5em} 
\subsubsection{Experimental Setting}
The MsFormer is trained with a batch size of 128 for 300 epochs. A leave-one-out strategy is applied to the NASA datasets, using three for training and one for testing. For both the C-MAPSS and NASA datasets, we employ a sliding window approach with a fixed window length of 28 to segment the original time series recordings. The ${Adam}$ optimizer is used with a learning rate of 0.001. All experiments are performed on an A100-SXM4-80GB GPU.
                           
\vspace{-1.5em}
\subsubsection{Evaluation Metrics}
The performance of the MsFormer model is evaluated using the root-mean-square error (RMSE), and the mean absolute error (MAE):
\begin{small}
\begin{align}
RMSE = \sqrt{\frac{1}{N} \sum_{i=1}^{n} (y_i - \hat{y}_i)^2}, \quad MAE = \frac{1}{N} \sum_{i=1}^{n} |y_i - \hat{y}_i|
\end{align}
\end{small}
where ${N}$ is the sample size, ${y_i}$ denotes the observed values, and ${\hat{y}_i}$ represents the model's predicted values. Additionally, we adopt the scoring function, which imposes a higher penalty on late predictions compared to RMSE:
\begin{small}
\begin{align}
Score = \begin{cases}
e^{\frac{y_i - \hat{y}_i}{13}} - 1, &  \hat{y}_i < y_i \\
e^{\frac{\hat{y}_i - y_i}{10}} - 1, &  \hat{y}_i \geq y_i
\end{cases}
\end{align}
\end{small}

\setlength{\tabcolsep}{3pt}
\begin{table*}[t]
  \centering
  \scriptsize
   \caption{Performance comparison of other methods on the C-MAPSS dataset.}
    \begin{tabular}{ccccccccc}
    \toprule
    Dataset & \multicolumn{2}{c}{FD001} & \multicolumn{2}{c}{FD002} & \multicolumn{2}{c}{FD003} & \multicolumn{2}{c}{FD004} \\
    \cmidrule(lr){2-3} \cmidrule(lr){4-5} \cmidrule(lr){6-7} \cmidrule(lr){8-9}
    Evaluation & RMSE & Score & RMSE & Score & RMSE & Score & RMSE & Score \\
    \midrule
    BiLSTM \cite{wang2018remaining} & 13.65 & 295.00 & 23.18 & 4130.00 & 13.74 & 317.00 & 24.86 & 5430.00 \\
    KDnet \cite{xu2021kdnet} & 13.68 & 362.08 & 14.47 & 929.20 & 12.95 & 327.27 & 15.96 & 1303.19 \\
    PE-Net \cite{jin2022position} & 13.98 & 280.87 & 14.69 & 881.73 & 12.33 & 272.85 & 15.40 & \underline{1103.18} \\
    TF-SCN \cite{jing2022transformer} & 12.05 & 219.00 & 14.71 & 1358.00 & 12.11 & 238.00 & 16.95 & 1367.00 \\
    IMDSSN \cite{zhang2023integrated} & 12.14 & 206.11 & 17.40 & 1775.15 & 12.35 & 229.54 & 19.78 & 2852.81  \\
    EAPN \cite{ZhangEAPN} & 12.11 & 245.32 & 15.68 & 1126.49 & 12.52 & 266.69 & 18.12 & 2050.72 \\
    GAT-DAT \cite{liang2023remaining} & 13.83 & 318.60 & 14.81 & 1163.80 & 14.85 & 438.50 & 16.80 & 1928.60\\
    DVGTformer \cite{wang2024dvgtformer} & \underline{11.33} & \underline{179.75} & \underline{14.28} & \underline{797.26} & 11.89 & 254.55 & 15.50 & 1107.50\\
    DTW-GPR \cite{zhou2024adaptive} & 12.02 & 212.59 & 17.38 & 1714.30 & 12.70 & 394.25 & 17.56 & 2509.30\\
    NSD-TGTN \cite{gao2024nonlinear} & 12.13 & 226.00 & 15.87 & 1477.00 & 12.01 & \underline{220.00} & 16.64 & 1493.00\\
    MLEAN \cite{liu2024multi} & 11.48 & 186.00 & 14.74 & 914.00 & 11.73 & 250.00 & 16.89 & 1370.00\\
    TFSCL \cite{jiang2025time} & 17.48 & 584.51 & 25.05 & 2852.54 & 16.38 & 1180.11 & 24.56 & 13899.48\\
    MSAN \cite{gao2025multiscale} & 11.42 & 198.00 & 14.91 & 1259.00 & \textbf{11.40} & \textbf{203.00} & \underline{15.13} & 1179.00\\
    \midrule
    MsFormer(\textit{ours}) & \textbf{10.94} & \textbf{178.50} & \textbf{13.64} & \textbf{766.70} & \underline{11.62} & 236.91 & \textbf{14.81} & \textbf{961.42}\\
    \bottomrule
    \end{tabular}%
  \label{tab:comp_methods}%
  \vspace{-2.0em} 
\end{table*}%

\vspace{-1.5em} 
\subsection{Comparison with Baselines}
\subsubsection{Performance on the C-MAPSS Dataset}
MsFormer performs satisfactorily on all subsets as shown in Table \ref{tab:comp_methods}. Notably, on the more complex FD002 and FD004 datasets, it surpasses all baselines with RMSEs/Scores of 13.64/766.70 and 14.81/961.42, respectively, demonstrating its superior capability in capturing intricate latent patterns across diverse temporal spans. MsFormer slightly underperforms compared to MSAN \cite{gao2025multiscale} and NSD-TGTN \cite{gao2024nonlinear} on FD003. MSAN uses an adaptive spatiotemporal feature extraction module, while NSD-TGTN employs a temporal graph transformer to mine long-term and spatial dependencies. However, their effectiveness diminishes on more complex datasets, with Scores of 1259/1477 on FD002 and 1179/1493 on FD004. This may be attributed to the reliance on spatial features and fixed-range temporal patterns, neglecting the multi-scale temporal correlations driven by machine working principles.

GAT-DAT \cite{liang2023remaining} and DVGTformer \cite{wang2024dvgtformer} capture time series correlations from the perspective of graph networks modeling, with DVGTformer performing competitively on FD001 and FD002 but still falling short of MsFormer. MLEAN \cite{liu2024multi} focuses on multi-scale and multi-frequency feature learning via a multi-channel attention network, and TFSCL \cite{jiang2025time} employs time-frequency synchronization contrastive learning. However, their performance is even inferior to traditional methods based on CNN (PE-Net \cite{jin2022position}) and GAN (KDnet \cite{xu2021kdnet}) frameworks on FD002 and FD004. This suggests that stacking self-attention layers without lightweight designs fails to guarantee reliable QoS on smaller-scale industrial datasets.

\begin{table}[htbp]
\vspace{-1.5em}
  \centering
  \scriptsize
  \caption{Performance comparison on the NASA Dataset, with values scaled by $10^2$.}
    \begin{tabular}{ccccccccccc}
    \toprule
    Dataset & \multicolumn{2}{c}{B5} & \multicolumn{2}{c}{B6} & \multicolumn{2}{c}{B7} & \multicolumn{2}{c}{B18} & \multicolumn{2}{c}{Average} \\
    \cmidrule(lr){2-3} \cmidrule(lr){4-5} \cmidrule(lr){6-7} \cmidrule(lr){8-9} \cmidrule(lr){10-11}
    Evaluation & RMSE & MAE & RMSE & MAE & RMSE & MAE & RMSE & MAE & RMSE & MAE \\
    \midrule
    DeTransformer \cite{chen2022transformer} & 6.08 & 5.26 & 14.84 & 13.67 & \textbf{3.86} & \underline{3.44} & 9.99 & 8.13 & 8.69 & 7.63 \\
    EM-UPF-W \cite{zhang2022remaining} & \underline{5.65} & \underline{4.58} & 10.10 & \underline{7.51} & 7.06 & 5.21 & 8.69 & 6.38 & 7.88 & 5.92 \\
    IMDSSN \cite{zhang2023integrated} & 8.54 & 7.36 & 12.03 & 10.67 & 9.46 & 8.51 & 5.65 & 4.63 & 8.92 & 7.79 \\
    Bao et al. \cite{bao2023multi} & 7.89 & 6.80 & 14.13 & 11.30 & 4.30 & 3.67 & 13.18 & 11.01 & 9.87 & 8.19\\
    Dual-Mixer \cite{fu2024supervised} & 7.90 & 7.20 & 11.17 & 10.15 & 7.75 & 6.12 & 6.30 & 5.47 & 8.28 & 7.23\\
    AttMoE \cite{chen2024attmoe} & 5.98 & 5.09 & 13.80 & 12.42 & 11.21 & 10.86 & 4.77 & 4.06 & 8.94 &8.11\\
    DSTN \cite{lin2024dual} & 5.87 & 4.70 & \underline{9.37} & 7.78 & 6.93 & 6.02 & \textbf{4.43} & \underline{3.65} & \underline{6.65} & \underline{5.54}\\
    \midrule
    MsFormer(\textit{ours}) & \textbf{4.93} & \textbf{3.80} & \textbf{5.34} & \textbf{4.13} & \underline{3.88} & \textbf{3.15} & \underline{4.49} & \textbf{3.32} & \textbf{4.66} & \textbf{3.60} \\
    \bottomrule
    \end{tabular}%
  \label{tab:comp_methods_battery}%
  \vspace{-4.0em}
\end{table}%

\vspace{-0.5em}
\subsubsection{Performance on the NASA Dataset}
To further evaluate the generalizability and robustness of MsFormer, we conduct additional experiments on the NASA Battery dataset, with results shown in Table \ref{tab:comp_methods_battery}. DeTransformer \cite{chen2022transformer} achieves a marginally lower RMSE of 3.86 compared with MsFormer on B7. Besides, AttMoE \cite{chen2024attmoe} combines the attention mechanism with a Mixture of Experts architecture, and DSTN \cite{lin2024dual} employs an autoencoder for feature extraction, followed by a transformer-based regressor for prediction. Although DSTN marginally outperforms MsFormer on subset B18 and achieves competitive results on B6, it still falls short of MsFormer's performance. These results highlight MsFormer's superior ability to model multi-scale temporal dependencies, substantiating its generalizability as a unified AI service model for reliable predictive maintenance.

\vspace{-1.0em}
\subsection{Ablation Studies}
\vspace{-0.2em}
\subsubsection{Multi-scale Sampling Module}
\label{subsubsec:DWP}

\begin{wraptable}{R}{0.45\textwidth} 
    \vspace{-3.5em} 
    \centering
    \scriptsize
    \caption{Results of the MS module applied at different stages on FD002.}
    \begin{tabular}{ccccc}
    \hline
    Stage 1  & Stage 2  & Stage 3  & RMSE & Score \\
    \midrule
      &   &    & 13.98 & 897.28 \\
      \text{\checkmark} &   &    & 14.07 & 876.18 \\
     &  \text{\checkmark} &   & 14.13 & 885.58 \\
     &  & \text{\checkmark} & 13.91 & 879.78 \\
    \text{\checkmark} & \text{\checkmark} &  & 14.03 & 848.40 \\
    \text{\checkmark} &  & \text{\checkmark} & 14.12 & \underline{825.45} \\
     & \text{\checkmark} & \text{\checkmark} & \underline{13.66} & 835.78 \\
     \midrule
    \text{\checkmark} & \text{\checkmark} & \text{\checkmark} & \textbf{13.64} & \textbf{766.70} \\
    \hline
    \end{tabular}
    \vspace{-2.5em} 
    \label{table:analysis_dwp}
\end{wraptable}

We first evaluate the effectiveness of the MS module by incrementally integrating it into each stage of MsFormer. As the down-sampled sequence length of MS increases progressively and is restored to the original sub-sequence length in the final stage, Table \ref{table:analysis_dwp} presents results for the first three stages. The absence of the MS module in all stages results in the lowest performance in terms of Score. Adding it to any single stage yields minimal improvement, while deploying it across two stages enhances performance but may cause semantic conflicts in subsequent layers if added randomly. Incorporating the MS module in all stages further boosts performance significantly, indicating its critical role in mitigating semantic sparsity and capturing temporal features from a global perspective.

\begin{figure}[htbp]
\vspace{-1.5em}
\centering
  \includegraphics[width=0.65\linewidth]{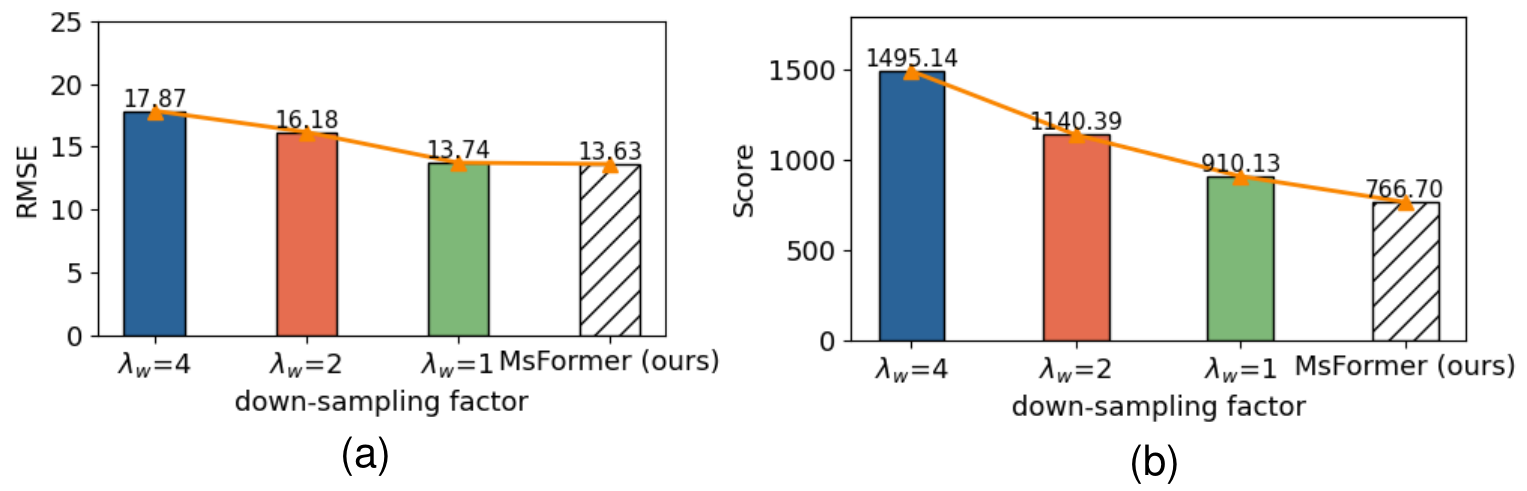}
   \vspace{-1.0em}
  \caption{Performance comparison with fixed dilation factor \( \lambda_w \) in the MS module.}
  \label{fig_window_length_fixed}
  \vspace{-2.0em}
\end{figure}

\begin{wrapfigure}{R}{0.45\textwidth}
    \vskip -2.0em
    \begin{minipage}{\linewidth}
    \centering 
    \includegraphics[width=\textwidth]{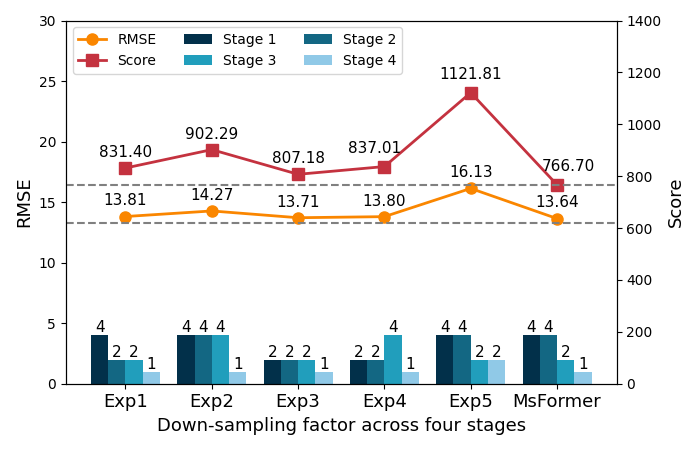} 
        \vskip -1.0em
    \caption{Results of various dilation factors \( \lambda_w \) in the MS module on FD002.}
    \label{fig_window_length_variable} 
        \vspace{-2.5em}
\end{minipage}
\end{wrapfigure}

The MS module adjusts the window length based on the down-sampling factor \( \lambda_w \). As shown in Figure \ref{fig_window_length_fixed}, a larger down-sampling factor corresponds to a wider temporal span between time points, which introduces semantic sparsity and hinders the model's ability to capture intrinsic temporal dependencies. Conversely, the proposed MsFormer model further optimizes performance through adaptive window-length expansion, achieving an RMSE/Score of 13.63/766.70. To further investigate the necessity of adaptive temporal resolution, we vary \( \lambda_w \) across four stages. As illustrated in Figure \ref{fig_window_length_variable}, gradually decreasing \( \lambda_w \) values achieves the best RUL prediction performance. This is because sparse sampling in the first two stages allows the LA blocks to capture temporal correlations, while denser sampling in the RPE block is crucial for finer temporal details. However, increasing \( \lambda_w \) at the last stage in Exp 5 significantly degrades performance, indicating the importance of preserving original position information. Fixed factor values at the first three stages in Exp 2 and Exp 3 restricts MsFormer's capacity to comprehensively capture temporal correlations, while inconsistent or larger settings in Exp 1 and Exp 4 impair performance by narrowing receptive fields or disrupting pattern learning.

\begin{table}[h]
  \centering
  \vspace{-1.5em}
  \scriptsize
  \caption{Results for position encoding mechanisms applied in the RPE block.}
    \begin{tabular}{ccccccccc}
    \toprule
    Dataset & \multicolumn{2}{c}{FD001} &\multicolumn{2}{c}{FD002} & \multicolumn{2}{c}{FD003} & \multicolumn{2}{c}{FD004}\\
    \cmidrule(lr){2-3} \cmidrule(lr){4-5} \cmidrule(lr){6-7} \cmidrule(lr){8-9}
    Evaluation & RMSE & Score & RMSE & Score & RMSE & Score & RMSE & Score \\
    \midrule
    w/o RPE & 12.31 & 212.10 & 14.06 & 941.34 & 12.60 & 277.92 & 15.00 & \underline{1021.74}\\
    Shaw et al. \cite{shaw-etal-2018-self} & 13.84 & 285.10 & 14.41 & 941.09 & 12.60 & 274.91 & 15.13 & 1004.25 \\
    DIET \cite{chen-etal-2021-simple} & \underline{10.99} & \textbf{169.45} & \textbf{13.57} & \underline{854.14} & 12.55 & \underline{263.21} & \underline{14.83} & 1037.11 \\
    T5’s RPE \cite{raffel2020exploring} & 11.87 & 194.43 & 14.04 & 887.36 & \underline{12.18} & 284.51 & \underline{14.83} & 1045.46 \\
    \cmidrule(lr){1-9}
    MsFormer(\textit{ours}) & \textbf{10.94} & \underline{178.50} &  \underline{13.64} & \textbf{766.70} & \textbf{11.62} & \textbf{236.91} & \textbf{14.81} & \textbf{961.42}  \\
    \bottomrule
    \end{tabular}
  \label{tab_analysis_rpe}
  \vspace{-4.0em}
\end{table}

\begin{wrapfigure}{R}{0.45\textwidth}
    \vskip -2.5em
    \begin{minipage}{\linewidth}
    \centering 
    \includegraphics[width=\textwidth]{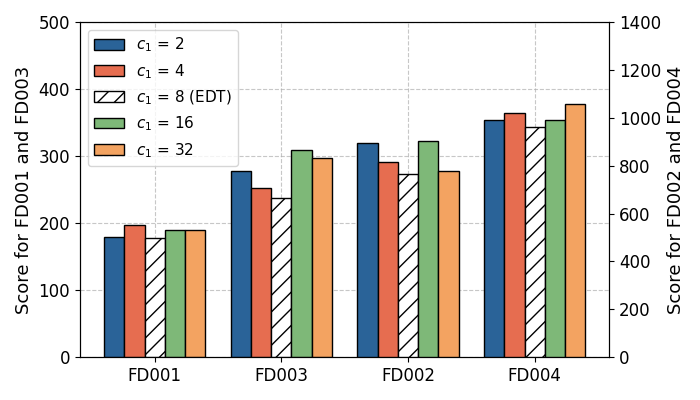} 
    \caption{Sensitivity analysis for parameter \(c_1\) in position encoding mechanisms.}
    \label{fig_analysis_rpe_c1} 
        \vspace{-2.0em}
\end{minipage}
\end{wrapfigure}

\subsubsection{Multi-scale Position Encoding Mechanism}
\label{subsubsec:rpe_c1}
We evaluate the effectiveness of the multi-scale position encoding mechanism in the RPE block. As shown in Table \ref{tab_analysis_rpe}, removing relative position encoding significantly degrades MsFormer's performance, underscoring its role in compensating for the positional information deficiency in self-attention and capturing spatial relationships in time series data. Unlike static position encoding methods such as Shaw et al. \cite{shaw-etal-2018-self}, DIET \cite{chen-etal-2021-simple}, and T5 \cite{raffel2020exploring}, which rely on fixed positional relationships, our multi-scale PE approach adapts weights within down-sampled sequences, enabling MsFormer to effectively integrate temporal information with element-wise positional relationships and achieve better generalization.

We further analyze the role of parameters in the proposed multi-scale position encoding mechanism, where parameters \(c_1\) and \(c_2\) serve as thresholds in the piecewise index function to measure the relative distance between elements. Considering the down-sampling window size from the MS module, \(c_1 = 8\) is chosen as the base value and \(c_2\)=\(2\times c_1\), with the sensitivity analysis illustrated in Figure \ref{fig_analysis_rpe_c1}. Reducing \(c_1\) to 4 limits attention to nearby elements, while increasing \(c_1\) to 16 assigns similar weights to both near and distant elements, leading to performance degradation particularly noticeable on FD002 and FD003. Extreme values (\(c_1 = 2\) or 32) render the piecewise function ineffective, either producing negligible positional encoding or adversely affecting model performance.

\vspace{-1.5em}
\begin{table}[h]
\centering
\scriptsize
\caption{Different attention layer setting strategies across stages of MsFormer. We diversely add attention layers to the RPE (Relative Position Encoding) block or replace the the LA (Lightweight Attention) blocks with attention layers from the RPE block.}
\begin{tabular}{ccccccccc}
\hline
Attention & \multicolumn{4}{c}{Stages} & \multicolumn{4}{c}{Score}\\
\cmidrule(lr){2-5} \cmidrule(lr){6-9}
 Layers & 1 & 2 & 3 & 4 & FD001 & FD002 & FD003 & FD004 \\
\midrule
 4 & LA & LA & RPE(2) & RPE(2) & 186.46 & 867.68 & \underline{309.12} & 1017.78\\
 4 & LA & LA & RPE(3) & RPE & \textbf{173.13} & 874.85 & 345.00 & 1040.54\\
 4 & LA & RPE & RPE(2) & PE & 258.08 & \underline{847.94} & 309.22 & \textbf{935.33}\\
 5 & RPE & RPE & RPE(2) & PE & 229.77 & 922.54 & 319.53 & 974.72\\
 \midrule
 3(\textit{ours}) & LA & LA & RPE(2) & RPE & \underline{178.50} & \textbf{766.70} & \textbf{236.91} & \underline{961.42}\\
\hline
\end{tabular}
\label{table:analysis_attention}
\vspace{-2.0em}
\end{table}

\subsubsection{Model Complexity}
\label{subsubsec: complexity}

To validate the lightweight design of MsFormer, we substitute the LA block with the RPE block in the first two stages and increase the number of attention layers in the RPE block. As shown in Table \ref{table:analysis_attention}, adding attention layers in the last two stages results in significant performance degradation on FD004, indicating that merely adding attention layers increases computational complexity without improving performance for industrial time series data. Replacing the LA block at the second stage slightly improves FD004 performance, likely due to its more intricate temporal dependencies. However, replacing all LA blocks with RPE blocks leads to performance deterioration. In conclusion, simply stacking self-attention mechanisms does not necessarily improve time-to-failure prediction performance. In contrast, MsFormer achieves superior performance while maintaining a lightweight architecture.

\begin{wraptable}{R}{0.45\textwidth} 
    \vspace{-2.5em} 
    \centering
    \small
    \caption{Complexity comparison.}
    \begin{tabular}{cc}
    \hline
    Evaluation & Params(M)\\
    \midrule
    FEDformer & \textgreater 114  \\
    Informer & \textgreater 11  \\
    Autoformer & \textgreater 10\\
    Pyraformer  & \textgreater 3\\ 
    Metaformer & \textgreater 12\\
    Swin Transformer & \textgreater 28\\
     \midrule
    MsFormer(\textit{all RPE}) &  1.05\\
    MsFormer(\textit{ours}) & \textbf{0.66}  \\
    \hline
    \end{tabular}
    \vspace{-3.0em} 
    \label{tab:analysis_complexity}
\end{wraptable}

We further compare the parameter count of MsFormer with popular Transformer-based models in Table \ref{tab:analysis_complexity}. In general time series forecasting, FEDformer \cite{zhou2022fedformer}, Informer \cite{zhou2021informer}, Autoformer \cite{wu2021autoformer}, and Pyraformer \cite{liu2022pyraformer} optimize the self-attention mechanism for enhanced performance. Dedicated to solving CV tasks, Metaformer \cite{yu2022metaformer} emphasizes token mixer flexibility while Swin Transformer \cite{liu2021swin} boosts efficiency with shifted windows in attention layers. The proposed MsFormer, achieves much lower complexity while maintaining superior performance for industrial prediction tasks.

\vspace{-1.0em}
\subsection{Visualization of Time-to-Failure Prediction Results}
\begin{figure}[htbp]
  \centering
  \includegraphics[width=0.7\linewidth]{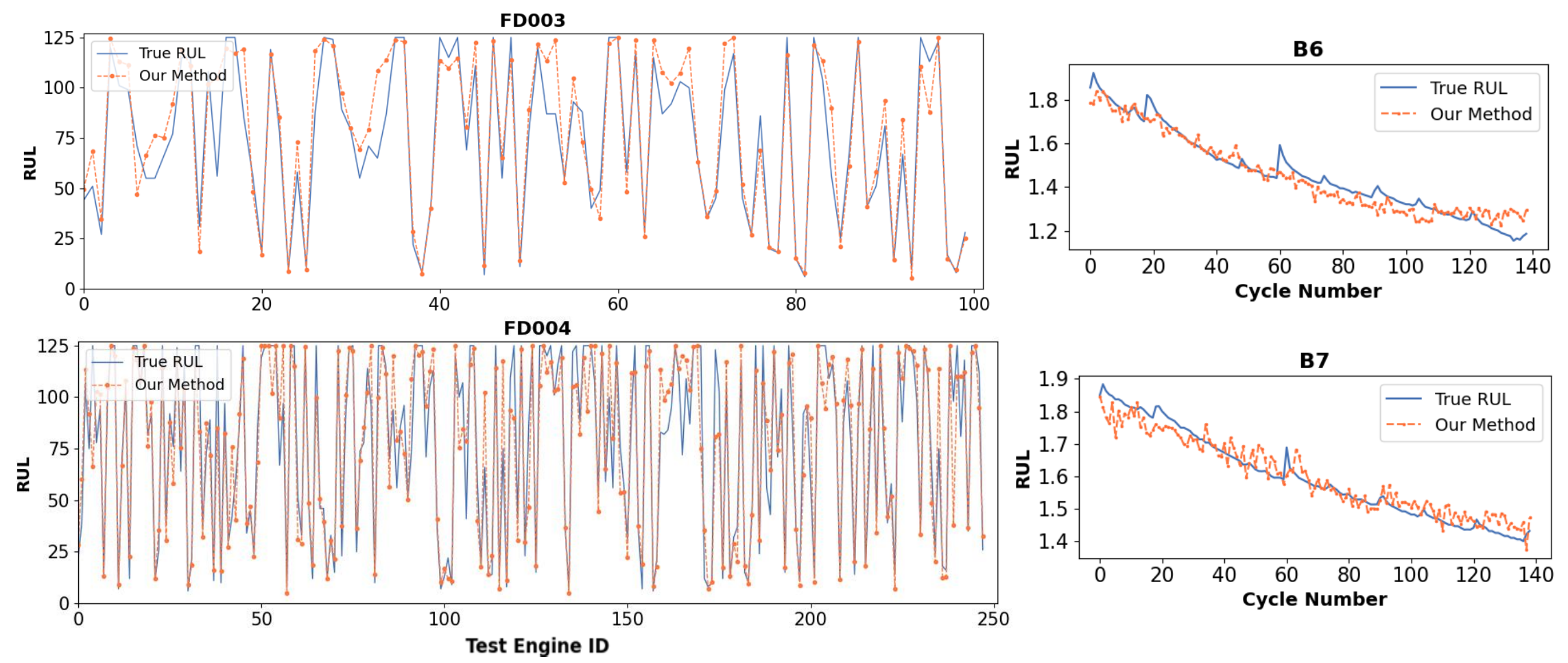}
  \caption{Visualization of time-to-failure prediction on C-MAPSS and NASA Datasets.}
  \label{fig_visual}
\vspace{-2.0em}
\end{figure}

\vspace{-0.5em}
The visualization results of the time-to-failure prediction task for MsFormer are illustrated in Figure \ref{fig_visual}. Regardless of the conditions of datasets, the prediction results by MsFormer can closely match the actual RUL values, particularly when the RUL values are low. This demonstrates that the MsFormer can accurately predict RUL, providing reliable support for the RUL warnings of mechanical devices. Consistent performance on NASA battery datasets further validates its robustness and generalizability as a lightweight unified AI service model for industrial predictive maintenance.

\vspace{-1.0em}

\section{Conclusion}
\label{sec:con}
\vspace{-0.5em}
In this paper, we propose MsFormer, a lightweight multi-scale Transformer designed as a unified AI service model for reliable industrial predictive maintenance. The model features a Multi-scale Sampling module to capture intricate temporal correlations in semantic-sparse IoT sensor data, and a multi-scale position encoding mechanism that integrates the down-sampling factor to describe the positional relationships within down-sampled sequences. Extensive experiments on real-world benchmarks across diverse devices and operating conditions validate the performance and strong generalizability of MsFormer as a robust predictive service. Beyond time-to-failure prediction, its lightweight and modular design makes MsFormer readily adaptable to broader industrial time series forecasting tasks with minimal modification, underscoring its practical value for real-world AIaaS deployments.



\vspace{-0.5em}
\begin{credits}
\subsubsection{\discintname}
The authors have no competing interests to declare that are relevant to the content of this article.
\end{credits}
\vspace{-1.0em}

%
%
%
\bibliographystyle{splncs04}
\bibliography{references}

@article{jin2022position,
  author={Jin, Ruibing and Wu, Min and Wu, Keyu and Gao, Kaizhou and Chen, Zhenghua and Li, Xiaoli},
  journal={IEEE/CAA Journal of Automatica Sinica}, 
  title={Position Encoding Based Convolutional Neural Networks for Machine Remaining Useful Life Prediction}, 
  year={2022},
  volume={9},
  number={8},
  pages={1427-1439}
}

@inproceedings{pascanu2013difficulty,
    author = {Pascanu, Razvan and Mikolov, Tomas and Bengio, Yoshua},
    title = {On the difficulty of training recurrent neural networks},
    year = {2013},
    publisher = {JMLR.org},
    booktitle = {Proceedings of the 30th International Conference on International Conference on Machine Learning - Volume 28},
    series = {ICML'13}
}

@inproceedings{zhou2022fedformer,
  title={{FEDformer}: Frequency enhanced decomposed transformer for long-term series forecasting},
  author={Zhou, Tian and Ma, Ziqing and Wen, Qingsong and Wang, Xue and Sun, Liang and Jin, Rong},
  booktitle={Proc. 39th International Conference on Machine Learning (ICML 2022)},
  year={2022}
}

@inproceedings{wu2021autoformer,
    author = {Wu, Haixu and Xu, Jiehui and Wang, Jianmin and Long, Mingsheng},
    title = {Autoformer: decomposition transformers with auto-correlation for long-term series forecasting},
    year = {2021},
    booktitle = {Proceedings of the 35th International Conference on Neural Information Processing Systems},
    numpages = {12},
    series = {NIPS '21}
}

@inproceedings{liu2024itransformer,
  title={iTransformer: Inverted Transformers Are Effective for Time Series Forecasting},
  author={Liu, Yong and Hu, Tengge and Zhang, Haoran and Wu, Haixu and Wang, Shiyu and Ma, Lintao and Long, Mingsheng},
  booktitle={The Twelfth International Conference on Learning Representations},
  year={2024}
}

@inproceedings{liu2022pyraformer,
title={Pyraformer: Low-Complexity Pyramidal Attention for Long-Range Time Series Modeling and Forecasting},
author={Liu, Shizhan and Yu, Hang and Liao, Cong and Li, Jianguo and Lin, Weiyao and Liu, Alex X and Dustdar, Schahram},
booktitle={International Conference on Learning Representations},
year={2022}
}

@article{liu2024multi,
  author={Liu, Xuezhen and Chen, Yongyi and Zhang, Dan and Yan, Ruqiang and Ni, Hongjie},
  journal={IEEE Transactions on Artificial Intelligence}, 
  title={A Multichannel Long-Term External Attention Network for Aeroengine Remaining Useful Life Prediction}, 
  year={2024},
  volume={5},
  number={10},
  pages={5130-5140}
}

@article{jiang2025time,
    author = {Li Jiang and Miaojun Wang and Peijie You and Xin Zhang and},
    title = {Time-frequency synchronisation contrastive learning-driven multi-sensor remaining useful life prediction},
    journal = {Nondestructive Testing and Evaluation},
    year = {2025},
    publisher = {Taylor \& Francis}
}

@article{zhang2024trend,
    title = {Trend-augmented and temporal-featured Transformer network with multi-sensor signals for remaining useful life prediction},
    journal = {Reliability Engineering \& System Safety},
    volume = {241},
    pages = {109662},
    year = {2024},
    author = {Yuru Zhang and Chun Su and Jiajun Wu and Hao Liu and Mingjiang Xie},
}

@article{gao2023dual,
  author={Gao, Hui and Li, Yibin and Zhao, Ying and Song, Yan},
  journal={IEEE Sensors Journal}, 
  title={Dual Channel Feature Attention-Based Approach for RUL Prediction Considering the Spatiotemporal Difference of Multisensor Data}, 
  year={2023},
  volume={23},
  number={8},
  pages={8514-8525}
}

@article{liang2023remaining,
    title = {Remaining useful life prediction via a deep adaptive transformer framework enhanced by graph attention network},
    journal = {International Journal of Fatigue},
    volume = {174},
    pages = {107722},
    year = {2023},
    author = {Pengfei Liang and Ying Li and Bin Wang and Xiaoming Yuan and Lijie Zhang}
}

@ARTICLE{gao2025multiscale,
    author={Gao, Zhan and Jiang, Weixiong and Wu, Jun and Dai, Tianjiao},
    journal={IEEE Sensors Journal}, 
    title={Multiscale Spatiotemporal Attention Network for Remaining Useful Life Prediction of Mechanical Systems}, 
    year={2025},
    volume={25},
    number={4},
    pages={6825-6835}
}

@article{jing2022transformer,
    title = {Transformer-based hierarchical latent space VAE for interpretable remaining useful life prediction},
    journal = {Advanced Engineering Informatics},
    volume = {54},
    pages = {101781},
    year = {2022},
    author = {Tao Jing and Pai Zheng and Liqiao Xia and Tianyuan Liu}
}

@article{gao2024nonlinear,
    title = {Nonlinear slow-varying dynamics-assisted temporal graph transformer network for remaining useful life prediction},
    journal = {Reliability Engineering \& System Safety},
    volume = {248},
    pages = {110162},
    year = {2024},
    author = {Zhan Gao and Weixiong Jiang and Jun Wu and Tianjiao Dai and Haiping Zhu},
}

@article{wang2024dvgtformer,
    title = {DVGTformer: A dual-view graph Transformer to fuse multi-sensor signals for remaining useful life prediction},
    journal = {Mechanical Systems and Signal Processing},
    volume = {207},
    pages = {110935},
    year = {2024},
    author = {Lei Wang and Hongrui Cao and Zhisheng Ye and Hao Xu and Jiaxiang Yan},
}

@article{zhang2024dual,
    title = {A dual-stream spatio-temporal fusion network with multi-sensor signals for remaining useful life prediction},
    journal = {Journal of Manufacturing Systems},
    volume = {76},
    pages = {43-58},
    year = {2024},
    author = {Qiang Zhang and Peixuan Yang and Qiong Liu},
}

@inproceedings{fang2024pbmt,
  author={Fang, Xing and Xiao, Lei and Shan, Yibing},
  booktitle={2024 Global Reliability and Prognostics and Health Management Conference (PHM-Beijing)}, 
  title={PBMT: A Novel Transformer-Based Model for Accurate RUL Prediction in Industrial Systems}, 
  year={2024}
}

@article{zhang2023integrated,
    title = {An integrated multi-head dual sparse self-attention network for remaining useful life prediction},
    journal = {Reliability Engineering \& System Safety},
    volume = {233},
    pages = {109096},
    year = {2023},
    author = {Jiusi Zhang and Xiang Li and Jilun Tian and Hao Luo and Shen Yin}
}

@article{chen2022transformer,
  author={Chen, Daoquan and Hong, Weicong and Zhou, Xiuze},
  journal={IEEE Access}, 
  title={Transformer Network for Remaining Useful Life Prediction of Lithium-Ion Batteries}, 
  year={2022},
  volume={10},
  pages={19621-19628}
}

@inproceedings{saxena2008damage,
  author={Saxena, Abhinav and Goebel, Kai and Simon, Don and Eklund, Neil},
  booktitle={2008 International Conference on Prognostics and Health Management}, 
  title={Damage propagation modeling for aircraft engine run-to-failure simulation}, 
  year={2008},
}

@inproceedings{wang2018remaining,
  author={Wang, Jiujian and Wen, Guilin and Yang, Shaopu and Liu, Yongqiang},
  booktitle={2018 Prognostics and System Health Management Conference (PHM-Chongqing)}, 
  title={Remaining Useful Life Estimation in Prognostics Using Deep Bidirectional LSTM Neural Network}, 
  year={2018},
  pages={1037-1042}
}

@article{xu2021kdnet,
  author={Xu, Qing and Chen, Zhenghua and Wu, Keyu and Wang, Chao and Wu, Min and Li, Xiaoli},
  journal={IEEE Transactions on Industrial Electronics}, 
  title={KDnet-RUL: A Knowledge Distillation Framework to Compress Deep Neural Networks for Machine Remaining Useful Life Prediction}, 
  year={2022}
}

@article{zhou2024adaptive,
    title = {An adaptive remaining useful life prediction model for aeroengine based on multi-angle similarity},
    journal = {Measurement},
    year = {2024},
    author = {Zhihao Zhou and Mingliang Bai and Zhenhua Long and Jinfu Liu and Daren Yu}
}

@article{zhang2022remaining,
  author={Zhang, Jiusi and Jiang, Yuchen and Li, Xiang and Luo, Hao and Yin, Shen and Kaynak, Okyay},
  journal={IEEE/ASME Transactions on Mechatronics}, 
  title={Remaining Useful Life Prediction of Lithium-Ion Battery With Adaptive Noise Estimation and Capacity Regeneration Detection}, 
  year={2023},
  pages={632-643}
}

@article{bao2023multi,
    author = {Bao, Qihao and Qin, Wenhu and Yun, Zhonghua},
    year = {2023},
    pages = {224},
    title = {A Multi-Stage Adaptive Method for Remaining Useful Life Prediction of Lithium-Ion Batteries Based on Swarm Intelligence Optimization},
    volume = {9},
    journal = {Batteries}
}

@article{fu2024supervised,
    title = {Supervised contrastive learning based dual-mixer model for Remaining Useful Life prediction},
    journal = {Reliability Engineering \& System Safety},
    volume = {251},
    pages = {110398},
    year = {2024},
    author = {En Fu and Yanyan Hu and Kaixiang Peng and Yuxin Chu}
}

@article{chen2024attmoe,
    title = {AttMoE: Attention with Mixture of Experts for remaining useful life prediction of lithium-ion batteries},
    journal = {Journal of Energy Storage},
    volume = {84},
    pages = {110780},
    year = {2024},
    author = {Daoquan Chen and Xiuze Zhou},
}

@article{lin2024dual,
  author = {Ching-Sheng Lin},
  title = {Dual Siamese transformer-encoder-based network for remaining useful life prediction},
  journal = {The Journal of Supercomputing},
  year = {2024},
  volume = {80},
  number = {17},
  pages = {25424--25449}
}

@inproceedings{shaw-etal-2018-self,
    title = {Self-Attention with Relative Position Representations},
    author = {Shaw, Peter and Uszkoreit, Jakob and Vaswani, Ashish},
    booktitle = {Proceedings of the 2018 Conference of the North {A}merican Chapter of the Association for Computational Linguistics: Human Language Technologies, Volume 2 (Short Papers)},
    year = {2018},
    pages = {464--468}
}

@inproceedings{chen-etal-2021-simple,
    title = {A Simple and Effective Positional Encoding for Transformers},
    author = {Chen, Pu-Chin and Tsai, Henry and Bhojanapalli, Srinadh  and Chung, Hyung Won and Chang, Yin-Wen and Ferng, Chun-Sung},
    booktitle = {Proceedings of the 2021 Conference on Empirical Methods in Natural Language Processing},
    year = {2021}
}

@article{raffel2020exploring,
    author = {Raffel, Colin and Shazeer, Noam and Roberts, Adam and Lee, Katherine and Narang, Sharan and Matena, Michael and Zhou, Yanqi and Li, Wei and Liu, Peter J.},
    title = {Exploring the limits of transfer learning with a unified text-to-text transformer},
    year = {2020},
    volume = {21},
    issn = {1532-4435},
    journal = {Journal of Machine Learning Research}
}

@inproceedings{zhou2021informer,
  author = {Haoyi Zhou and Shanghang Zhang and Jieqi Peng and Shuai Zhang and Jianxin Li and Hui Xiong and Wancai Zhang},
  title = {Informer: Beyond Efficient Transformer for Long Sequence Time-Series Forecasting},
  booktitle = {The Thirty-Fifth {AAAI} Conference on Artificial Intelligence, {AAAI} 2021, Virtual Conference},
  volume    = {35},
  pages     = {11106--11115},
  year = {2021}
}

@inproceedings{yu2022metaformer,
  author={Yu, Weihao and Luo, Mi and Zhou, Pan and Si, Chenyang and Zhou, Yichen and Wang, Xinchao and Feng, Jiashi and Yan, Shuicheng},
  booktitle={2022 IEEE/CVF Conference on Computer Vision and Pattern Recognition (CVPR)}, 
  title={MetaFormer is Actually What You Need for Vision}, 
  year={2022},
  pages={10809-10819}
}

@inproceedings{liu2021swin,
  title={Swin Transformer: Hierarchical Vision Transformer using Shifted Windows},
  author={Liu, Ze and Lin, Yutong and Cao, Yue and Hu, Han and Wei, Yixuan and Zhang, Zheng and Lin, Stephen and Guo, Baining},
  booktitle={Proceedings of the IEEE/CVF International Conference on Computer Vision (ICCV)},
  year={2021}
}

@article{ZhangEAPN,
title = {Predicting remaining useful life of a machine based on embedded attention parallel networks},
journal = {Mechanical Systems and Signal Processing},
volume = {192},
pages = {110221},
year = {2023},
author = {Xiong Zhang and Yunfei Guo and Hong Shangguan and Ranran Li and Xiaojia Wu and Anhong Wang}
}

@inproceedings{Zeng2022AreTE,
  title={Are Transformers Effective for Time Series Forecasting?},
  author={Ailing Zeng and Muxi Chen and Lei Zhang and Qiang Xu},
  booktitle={Proceedings of the AAAI Conference on Artificial Intelligence},
  year={2023}
}

@article{Robotics,
author = {Ali, Amir R. and Kamal, Hossam},
title = {Time-to-Fault Prediction Framework for Automated Manufacturing in Humanoid Robotics Using Deep Learning},
journal = {Technologies},
volume = {13},
year = {2025},
number = {2},
artical-number = {42},
ISSN = {2227-7080}
}

@inproceedings{nips_rnnvanishing,
 author = {Zucchet, Nicolas and Orvieto, Antonio},
 booktitle = {Advances in Neural Information Processing Systems},
 pages = {139402--139443},
 publisher = {Curran Associates, Inc.},
 title = {Recurrent neural networks: vanishing and exploding gradients are not the end of the story},
 volume = {37},
 year = {2024}
}

@inproceedings{Timexer,
 author = {Wang, Yuxuan and Wu, Haixu and Dong, Jiaxiang and Qin, Guo and Zhang, Haoran and Liu, Yong and Qiu, Yunzhong and Wang, Jianmin and Long, Mingsheng},
 booktitle = {Advances in Neural Information Processing Systems},
 editor = {A. Globerson and L. Mackey and D. Belgrave and A. Fan and U. Paquet and J. Tomczak and C. Zhang},
 pages = {469--498},
 publisher = {Curran Associates, Inc.},
 title = {TimeXer: Empowering Transformers for Time Series Forecasting with Exogenous Variables},
 volume = {37},
 year = {2024}
}

@inproceedings{wang2024timemixer,
title={TimeMixer: Decomposable Multiscale Mixing for Time Series Forecasting},
author={Shiyu Wang and Haixu Wu and Xiaoming Shi and Tengge Hu and Huakun Luo and Lintao Ma and James Y. Zhang and JUN ZHOU},
booktitle={The Twelfth International Conference on Learning Representations},
year={2024}
}

@misc{lu2024incontextpredictor,
      title={In-context Time Series Predictor}, 
      author={Jiecheng Lu and Yan Sun and Shihao Yang},
      year={2024},
      eprint={2405.14982},
      archivePrefix={arXiv},
      primaryClass={cs.LG}
}

@inproceedings{liu2025timerxl,
title={Timer-{XL}: Long-Context Transformers for Unified Time Series Forecasting},
author={Yong Liu and Guo Qin and Xiangdong Huang and Jianmin Wang and Mingsheng Long},
booktitle={The Thirteenth International Conference on Learning Representations},
year={2025}
}

@inproceedings{masserano2025enhancing,
title={Enhancing Foundation Models for Time Series Forecasting via Wavelet-based Tokenization},
author={Luca Masserano and Abdul Fatir Ansari and Boran Han and Xiyuan Zhang and Christos Faloutsos and Michael W. Mahoney and Andrew Gordon Wilson and Youngsuk Park and Syama Sundar Rangapuram and Danielle C. Maddix and Bernie Wang},
booktitle={Forty-second International Conference on Machine Learning},
year={2025}
}

\end{document}